\pdfoutput=1

\documentclass[11pt]{article}

\usepackage[]{latex/acl}

\usepackage{times}
\usepackage{latexsym}

\usepackage[T1]{fontenc}

\usepackage[utf8]{inputenc}

\usepackage{microtype}

\usepackage{inconsolata}

\usepackage{xspace}
\usepackage{enumitem} 
\usepackage{booktabs}
\usepackage[capitalize]{cleveref}
\usepackage{graphicx}
\usepackage{multirow}

\makeatletter
\DeclareRobustCommand\onedot{\futurelet\@let@token\@onedot}
\def\@onedot{\ifx\@let@token.\else.\null\fi\xspace}
\def\eg{\emph{e.g}\onedot} \def\Eg{\emph{E.g}\onedot}
\def\ie{\emph{i.e}\onedot}

\makeatother

\crefname{section}{Sec.}{Secs.}
\Crefname{section}{Section}{Sections}
\Crefname{table}{Table}{Tables}
\crefname{table}{Tab.}{Tabs.}

\title{Machine-Generated Text Localization}

\author{Zhongping Zhang \qquad Wenda Qin \qquad Bryan A. Plummer \\
  Boston University\\
  \texttt{ \{zpzhang, wdqin, bplum\}@bu.edu} 
  }

\begin{document}

\maketitle
\begin{abstract}

Machine-Generated Text (MGT) detection aims to identify a piece of text as machine or human written.  Prior work has primarily formulated MGT detection as a binary classification task over an entire document, with limited work exploring cases where only part of a document is machine generated.  This paper provides the first in-depth study of MGT that localizes the portions of a document that were machine generated.  Thus, if a bad actor were to change a key portion of a news article to spread misinformation, whole document MGT detection may fail since the vast majority is human written, but our approach can succeed due to its granular approach.  A key challenge in our MGT localization task is that short spans of text, \eg, a single sentence, provides little information indicating if it is machine generated due to its short length.  To address this, we leverage contextual information, where we predict whether multiple sentences are machine or human written at once.  This enables our approach to identify changes in style or content to boost performance.  A gain of 4-13\% mean Average Precision (mAP) over prior work demonstrates the effectiveness of approach on five diverse datasets: GoodNews, VisualNews, WikiText, Essay, and WP.  We release our implementation at \href{https://github.com/Zhongping-Zhang/MGT_Localization}{this http URL}.

\end{abstract}

\section{Introduction}
\label{sec:introduction}

\begin{figure}[t]
    \centering
    \includegraphics[width=1.0\linewidth]{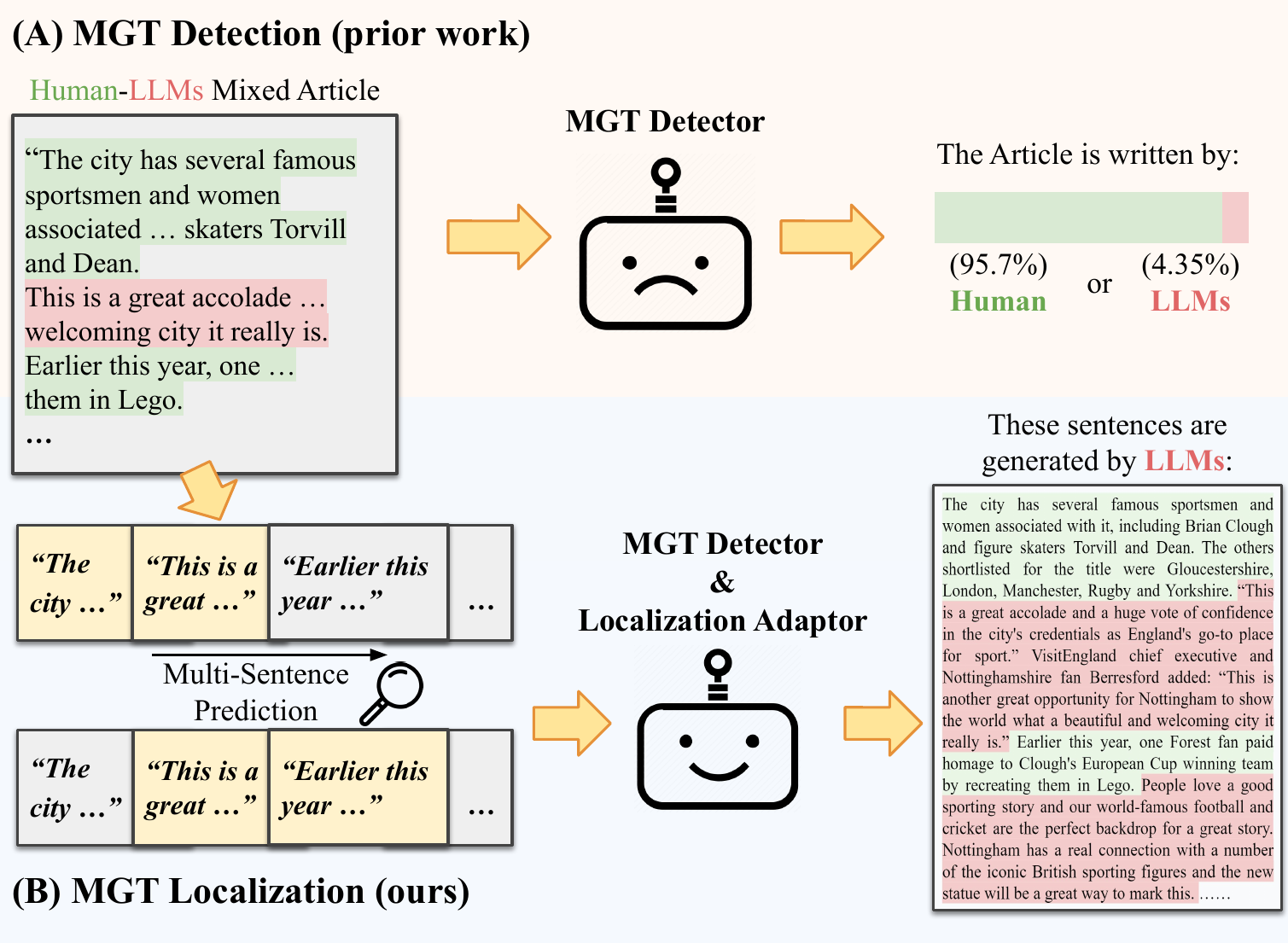}
    \caption{Prior work in machine-generated text detection~\cite{mitchell2023detectgpt, su2023detectllm, guo2023close}, shown in \textbf{(A)}, predicts a binary label indicating if an entire document or paragraph was machine or human generated. However, real-world articles may contain a mix of human-written and machine-generated sentences, which are challenging to detect when only a small part of the document was changed.  To address this, we explore machine-generated text localization, shown in \textbf{(B)}, where we introduce a lightweight localization adaptor to perform sentence-level predictions within a text document. Our method predicts multiple sentences at once to address challenges caused by the text's short length.
    }
    \label{fig:overview}
\end{figure}

Large Language Models (LLMs)~\cite{brown2020language, ouyang2022training, gpt-j, gao2020pile, du2022glm, touvron2023llama} have led to significant advancements in many domains like conversational systems~\cite{openai2023gpt}, social media data mining~\cite{lyu2023gpt}, and medical image analysis~\cite{nori2023capabilities}, among others. Many ethical or factual problems can arise in text generation such as hallucination~\cite{lin2022truthfulqa} or misuse for monetization (ad revenue through clicks) or propaganda~\cite{zellers2019defending}. Machine-generated-text (MGT) detection can help defend against this misuse. However, as shown in \Cref{fig:overview}(A), prior work has primarily focused on whole document (\ie, binary) classification as human or machine generated (\eg,~\citealp{tanDIDAN2020, mitchell2023detectgpt, verma2023ghostbuster, guo2023close, su2023detectllm}), but many applications may mix machine-generated and human-written text. For example, bad actors might use LLMs to manipulate certain sections of a news article to spread misinformation. Thus, whole document classification may fail since most text is human written. While \citet{verma2023ghostbuster} did explore paragraph-level detection, this may still be too coarse to detect changes to single sentences. We also note that a concurrent work, \citet{wang2023seqxgpt}, attempted to achieve sentence-level predictions. However, their approach primarily addresses articles where the initial segment is human-written and the second segment is AI-generated. In contrast, our paper considers a more challenging and general case in which multiple sections of an article can be generated by LLMs. A specific comparison is presented in \Cref{sec:appendix_seqxgpt}.


To bridge this gap, we introduce the first in-depth study on machine-generated text \emph{localization}. As illustrated in \Cref{fig:overview}(B), our task's goal is to identify any machine generated sentences within a given article.
A straightforward approach for our localization task would be simply employing sliding windows on top of existing detectors~\cite{solaiman2019release, ouyang2022training}. While this enables us to adapt existing binary classification methods to our task (\eg, OpenAI-Detector~\cite{solaiman2019release}, DetectGPT~\cite{wang2023detectgpt}, and ChatGPT-Detector~\cite{guo2023close}), these detectors perform poorly on the inherently short length of sentences (\ie, many of these models reported that reliable classification requires sentences to be longer than 50 tokens).

To address the aforementioned issues, we propose a lightweight Adaptor network for generated text Localization (AdaLoc).  Our approach provides additional context by including multiple sentences at once, but then predicts whether each individual sentence is machine generated. This way the model has more information when making its predictions, but still produces dense labels. We find our approach can be further improved by aggregating overlapping predictions using a majority vote. Our experiments show our approach outperforms direct adaptations of state-of-the-art on MGT detection, training a model for MGT localization without context (\ie, directly on the sentences it is trying to label), or aggregating overlapping predictions when a single label is produced over blocks of text.


In summary our contributions are:
\begin{itemize}[nosep,leftmargin=*]
    \item  We provide the first in-depth study on machine-generated text localization\footnote{We note that there are tools like GPT-Zero~\cite{tian2023gptzero} or Copyleaks~\cite{copyleaks} capable of performing sentence-level analysis on machine-generated text. However, since these tools haven't released any papers about how they identify these sentences, we believe our claim is warranted.}. This task bridges the gap between the current binary classification task and articles that contain a mix of human and machine-generated text.
    \item We introduce a data creation pipeline to generate articles consisting of both human-written and machine-generated texts. This approach can be used to automatically generate training and evaluation data for our MGT localization task.
    \item We identify a major challenge in machine-generated text localization arising from inaccurate judgements for short texts. To address this challenge, we use a majority vote strategy from overlapping predictions with our AdaLoc approach to provide dense labels over sentences in an article. 
    \item The effectiveness of our proposed methods are validated on five diverse datasets (GoodNews, VisualNew, WikiText, Essay, and WP), with a 4$\sim$13\% mAP improvement.
\end{itemize}

\section{Related Work}
\label{sec:related_work}



The importance of detecting machine-generated text has risen due to the risk of producing factual inaccuracies~\cite{lin2022truthfulqa} and the potential for its use in misinformation, such as propaganda or monetization~\cite{zhang2023show}. Existing detection methods can primarily be categorized into two types: metric-based methods and model-based methods. Metrics-base methods~\cite{solaiman2019release, gehrmann2019gltr, mitchell2023detectgpt, su2023detectllm, wang2023detectgpt} rely on extracting distinguishable features from text using the target language model. Specifically, \citet{solaiman2019release} apply log probability to identify whether a document is generated by LLMs or humans. \citet{gehrmann2019gltr} employ the absolute rank of each token as the evaluation metrics. Recent studies~\cite{mitchell2023detectgpt, su2023detectllm, bao2023fast} have shown that minor modifications to machine-generated text usually result in lower log probability under the model than the original text, a pattern not observed with human-written text. Thus, these methods introduce perturbations to the input text, measuring the discrepancy between the original and perturbed texts. 

Model-based methods~\cite{solaiman2019release, guo2023close, ippolito2020automatic, bhattacharjee2023conda} involve training specific classifiers on annotated corpora to classify input text directly. This kind of method is particular useful for detecting text generated by black-box or unknown models. For example, \citet{solaiman2019release} fine-tuned a RoBERTa~\cite{liu2019roberta} model based on outputs from GPT-series models. \citet{guo2023close} developed their approach using the HC3~\cite{guo2023close} dataset. 

To improve the generalization capabilities of these detectors, \citet{verma2023ghostbuster} extracted features from text using a series of language models and trained a classifier to categorize these features. However, all the methods we have discussed explored generated text detection at coarse scales (\ie, the whole document or paragraph level).  In contrast, our paper broadens the discussion to incorporate articles comprising both human-written and machine-generated content at a granular (sentence) level where prior work underperforms.

\section{Machine-generated Text Localization}
\label{sec:MGTL}
Given an article $x$ containing sentences $S = \{s_1,...,s_n\}$, Machine-Generated Text (MGT) localization aims at identifying specific sentences produced by LLMs. Unlike the MGT detection task that assigns a single label $y$ for the whole document, our task predicts a sequence of labels $\{y_1,...,y_n\}$, where each label $y_i$ corresponds to an individual sentence $s_i$, providing a more precise indicator of machine-generated content in $x$. 



A straightforward baseline to adapt existing methods~\cite{mitchell2023detectgpt, su2023detectllm, guo2023close} to the localization task is predicting labels sentence by sentence\footnote{Although single-sentence prediction can be directly applied here, it cannot leverage context due to its single-input single-output nature. In contrast, MGT localization predicts a sequence labels $\{y_1,...,y_n\}$ to identify AI-generated sentences given an article $S = \{s_1,...,s_n\}$, improving prediction of each label $y_i$ with context (\eg, $s_{i-2}$, $s_{i-1}$, $s_{i+1}$, $s_{i+2}$).} (sliding window). The major challenge here is that a single sentence often provides insufficient information to determine whether it is machine-generated due to its short length. To address this, we leverage the contextual information to improve performance, where our method predicts multiple sentences at once so that changes in style or content can be identified.

\begin{figure*}[t]
    \centering
    \includegraphics[width=1.0\linewidth]{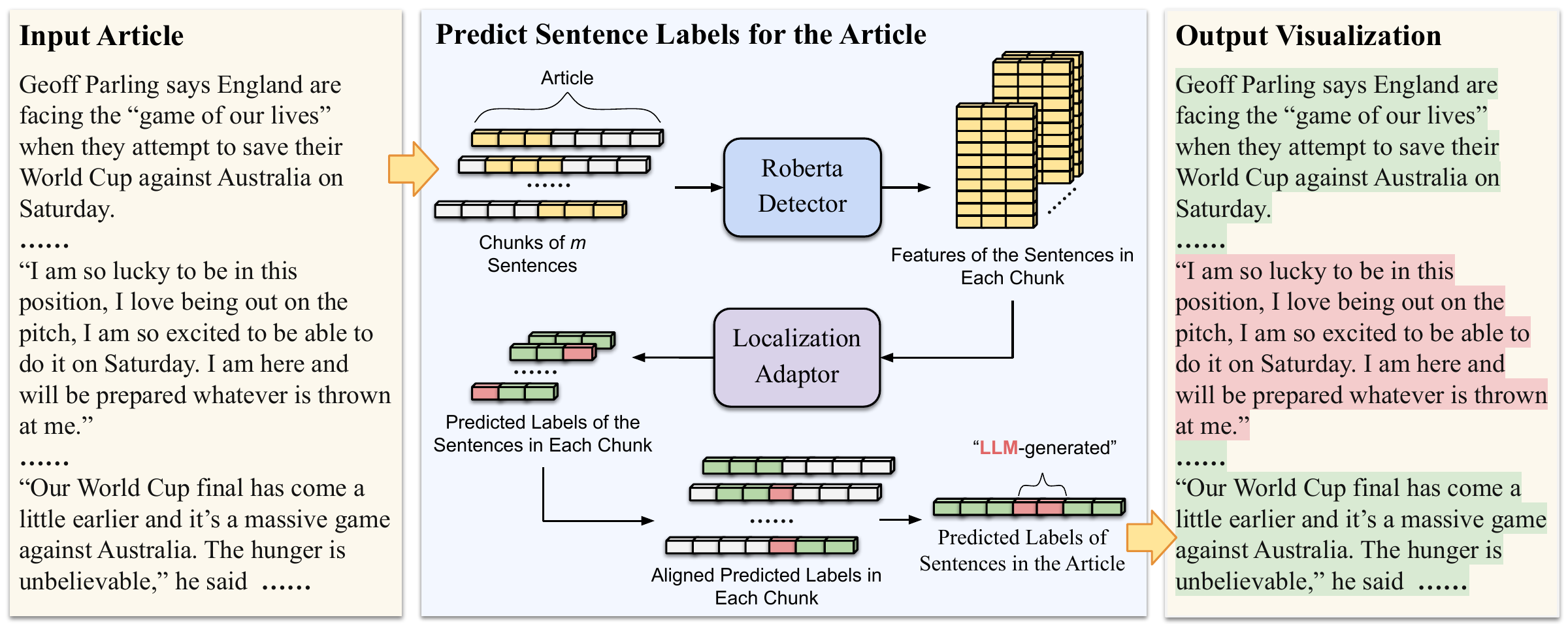}
    \caption{\textbf{Roberta+AdaLoc Overview.} Our method first divides an article into several chunks, each containing $m$ sentences. We then employ existing MGT detection methods (\eg, \citealp{solaiman2019release} or \citealp{guo2023close}) to extract chunk features. The model parameters in this phase are fixed, eliminating the need for further training. To assign a specific label to each sentence, we introduce a lightweight localization adaptor, AdaLoc. AdaLoc consists of two fully connected layers, with the output from the final layer being an $m\times1$ vector. This vector's elements represent the predicted labels for the sentences within the chunk. See \Cref{sec:MGTL_methods} for detailed information.}
    \label{fig:MGTL_framework}
\end{figure*}

Specifically, \Cref{sec:MGTL_data_preparation} introduces our method for constructing manipulated articles, which serve as the training and evaluation data for our experiments. \Cref{sec:MGTL_methods} present our methods to adapt established detectors to the MGT localization task. We first discuss a majority vote algorithm that predicts multiple sentences simultaneously to improve the single-sentence prediction. Given that this method assigns the same label to all sentences within a window, it requires a trade-off between the window size and the granularity of localization. In order to enhance performance without sacrificing granularity, we propose a lightweight adaptor designed to predict multiple sentence at once and allocate a unique label to each. \Cref{fig:MGTL_framework} provides an overview of this method.

\subsection{Data Preparation: Article Manipulation}
\label{sec:MGTL_data_preparation}
As discussed in the Introduction, real-world articles might contain a mix of human-written and machine-generated text. To prepare such articles for our training and evaluation datasets, we use LLMs to produce sentences conditioned on the title and initial paragraphs of each article. Then, we substitute certain sections (\eg, paragraphs or sentences) of the original article with these machine-generated sentences. Following \citet{mitchell2023detectgpt}, we use a variety of language models for text generation\footnote{For Essay and WP~\cite{verma2023ghostbuster} datasets, we directly combine the machine-generated text and human-written text to get such articles.}, including GPT2-1.5B~\cite{radford2019language}, GPTNeo-2.7B~\cite{gao2020pile}, GPTJ-6B~\cite{gpt-j}, OPT-2.7B~\cite{zhang2022opt}, and GPTNeoX-20B~\cite{gao2020pile}. During the generation process, we employ two sampling methods: top-$k$ sampling with $k$ set to 40, and top-$p$ sampling with $p$ ranging from 0.94 to 0.98. To maintain sentence integrity, we apply NLTK~\cite{bird2009natural} for segmenting complete paragraphs into sentences, selecting only well-formed sentences for inclusion in the articles. Each article is combined with 1$\sim$3 MGT segments, with segment lengths varying from 40 to 300 tokens.

\subsection{Methods}
\label{sec:MGTL_methods}

\noindent \textbf{Single-sentence Prediction.} To achieve sentence-level predictions for MGT, one straightforward approach is to employ a sliding window technique, applying it to each sentence within an article using established methods (\eg, \citealp{solaiman2019release} or \citealp{guo2023close}). This strategy allows for the generation of a sequence of labels throughout the article, pinpointing specific sentences that are machine-generated. A major challenge with this approach is the short length of the input text. Consistent with previous studies~\cite{solaiman2019release, mitchell2023detectgpt, verma2023ghostbuster}, we find that MGT detectors produce reliable outcomes with inputs exceeding 50 tokens, while individual sentences typically fall short of this token count. To address this limitation, we propose expanding the window size to incorporate multiple sentences at once, as detailed in the following paragraphs.
\smallskip

\noindent \textbf{Multi-sentence Prediction with Majority Vote.} In this approach, each window processes multiple sentences $\{s_1,...,s_m\}$ as input, assigning the same label to all sentences within that window. Consequently, when the window step is set to $1$, sentences within the same article are labeled $m$ times. The final label for each sentence $s_j$ is then determined based on these $m$ labels, using a majority vote approach. Due to the uniform labeling within each window, this method necessitates balancing the window size against the localization granularity. As our experiments will show, this strategy improves performance compared to single-sentence predictions with an appropriate number of sentences per window, validating that the challenge of short text detection can be mitigated by increasing the window size.
\smallskip

\noindent \textbf{Multi-sentence Prediction with Localization Adaptor.} 
To boost the classification precision without reducing localization granularity, we further propose a lightweight localization adaptor, AdaLoc, capable of predicting multiple sentences simultaneously and assigning them corresponding labels. As shown in \Cref{fig:MGTL_framework}, our method first divides the article into different chunks by NLTK~\cite{bird2009natural}, each comprising sentences $\{s_1,...,s_m\}$. Leveraging the pretrained Roberta model~\cite{solaiman2019release}, we obtain chunk features with dimensions of $512\times1024$. The [CLS] token's~\cite{liu2019roberta} chunk features are extracted for the input to AdaLoc. AdaLoc consists of two Fully Connected (FC) layers configured as 1024-1024-$m$, incorporating dropout after the first FC layer. The output of AdaLoc is a $m\times1$ vector $\{p_1,...,p_m\}$, with each element denoting the label of a corresponding sentence within the chunk (window). Given $m$ sentences, AdaLoc can predict $2^m$ potential binarization vectors. We apply binary cross entropy loss to finetune AdaLoc.

\section{Experiments}
\label{sec:experiments}

\begin{table*}[t]
\centering
\begin{tabular}{l ccccccc}
\toprule
\textbf{Model} & GPT-2 & GPT-Neo & OPT & GPT-J & GPT-NeoX & \textbf{mAP} & \textbf{All} \\
Scale & -1.5B & -2.7B & -2.7B & -6B & -20B \\
\midrule
\multicolumn{7}{c}{\textbf{AP on GoodNews~\cite{biten2019good}} }\\
\midrule
Random & 23.14 & 22.80 & 22.30 & 22.67 & 22.12 & 22.61 & 22.51 \\
All 0/1 & 23.09 & 22.73 & 22.11 & 22.87 & 22.30 & 22.62 & 22.54 \\
DetectGPT & 48.91 & 48.87 & 46.40 & 49.87 & 46.69 & 48.15 & 47.53 \\
DetectLLM & 50.04 & 47.66 & 47.56 & 48.51 & 47.18 & 48.19 & 47.79 \\
ChatGPT-D & 32.51 & 31.35 & 30.94 & 30.10 & 28.85 & 30.75 & 30.64 \\
Roberta-B & 45.02 & 44.97 & 39.96 & 38.99 & 35.13 & 40.81 & 40.74 \\
Roberta-L & 57.24 & 58.05 & 49.38 & 47.41 & 41.32 & 50.67 & 50.85 \\
\hline
Roberta-B+vote & 60.92 & 61.31 & 55.56 & 53.63 & 48.51 & 55.99 & 56.24 \\
Roberta-L+vote & 71.03 & 72.03 & 64.37 & 62.28 & 55.39 & 65.02 & 65.50 \\
Roberta-L+AdaLoc & \textbf{82.82} & \textbf{82.46} & \textbf{78.69} & \textbf{76.90} & \textbf{71.62} & \textbf{78.49} & \textbf{79.13} \\
\bottomrule
\end{tabular}
\caption{\textbf{Text Localization Results on GoodNews~\cite{biten2019good}.} \emph{vote} denotes multi-sentence prediction with majority vote, \emph{AdaLoc} denotes multi-sentence prediction with our localization adaptor. For both methods, the window spans 3 sentences. To make a fair comparison, AdaLoc is finetuned only on GPT-2 generated articles, with the same procedure for Roberta-L. We observe that both \emph{vote} and \emph{AdaLoc} notably enhance the localization precision compared to single-sentence prediction. See \Cref{sec:experiments_goodnews} for detailed discussion.}
\label{table:goodnews_localization}
\end{table*}

\subsection{Datasets \& Metrics}
\noindent \textbf{GoodNews}~\cite{biten2019good} is a news dataset that provides the URLs of New York Times articles from 2010 to 2018. After filtering out broken links and non-English articles, we randomly select 10,000, 1,000 and 1,000 articles for training, validation and test sets.
\smallskip

\noindent \textbf{VisualNews}~\cite{liu2020visualnews} contains articles from four news sources: \emph{Guardian}, \emph{BBC}, \emph{USA Today}, and \emph{Washington Post}. Similarly, we randomly select 1,000 articles for evaluation. Another 1,000 articles are used to train logistic classifiers for metric-based methods like DetectGPT~\cite{mitchell2023detectgpt} and DetectLLM~\cite{su2023detectllm}. 
\smallskip

\noindent \textbf{WikiText}~\cite{merity2016pointer} contains 600/60/60 Wikipedia articles in training/validation/test sets, respectively. We use the test set of WikiText directly for our evaluation.
\smallskip

\noindent \textbf{Essay \& WP}~\cite{verma2023ghostbuster} are designed for assessing AI-generated text detection in student essays and creative writings. Similar to the news articles, we randomly choose 1,000 human-authored documents from each dataset and blend them with AI-generated text (ChatGLM, ChatGPT, and GPT-4) for our analysis.
\smallskip

\noindent \textbf{Metrics.} Our experiments begin with the use of Average Precision (AP) to measure prediction accuracy for articles sampled from specific LLMs. We then compute the mean AP (mAP) based on documents generated by different LLMs. In addition, we aggregate predicted labels from all articles to calculate their collective AP (referred to as ``All'' in our comparisons). This metric allows us to evaluate a detector's performance across texts produced by various LLMs.

\subsection{Baselines}
\noindent \textbf{Data Bias.} We apply ``Random'' and ``All 0/1'' strategies to evaluate the data bias in our datasets.
\smallskip

\noindent \textbf{DetectGPT}~\cite{mitchell2023detectgpt} is a metric-based approach that introduces perturbations to the original text. This method is based on the intuition that LLM-derived text tends to be situated at the local optimum of the model's log probability function. Therefore, perturbations are likely to lower the log probability of machine-generated text, while the effect on human-written text is more variable.
\smallskip

\noindent \textbf{DetectLLM}~\cite{su2023detectllm} is a metric-based method which combines Log-Likelihood and Log-Rank (LRR) as its evaluation metric. For DetectGPT and DetectLLM, we trained a classifier for each domain's data distribution to determine the thresholds between machine-generated and human-written text.
\smallskip

\noindent \textbf{ChatGPT-D}~\cite{guo2023close} is proposed to detect texts generated by ChatGPT. This detector is trained on HC3~\cite{guo2023close} dataset, which consists of 40k questions and their answers, written by both humans and ChatGPT.
\smallskip

\noindent \textbf{Roberta-D}~\cite{solaiman2019release} is a model trained on the output of GPT2, released by OpenAI. It can be generalized to outputs from other LLMs by fine-tuning with early stopping.
\smallskip

\noindent \textbf{Roberta-MPU}~\cite{tian2024multiscale} is a framework designed to address the challenge of short-text detection. By incorporating a length-sensitive loss and a multiscale module, Roberta-MPU enhances the detection of short texts without compromising the performance on long texts.
\smallskip

\noindent \textbf{GPT-zero}~\cite{tian2023gptzero} is an online tool for analyzing whether a piece of text is human-written or machine-generated. We apply it as an external, ``blackbox'' model to assess its performance in our localization task.

\subsection{MGT Localization on GoodNews.}
\label{sec:experiments_goodnews}

\begin{figure*}[t!]
    \centering
    \includegraphics[width=1.0\linewidth]{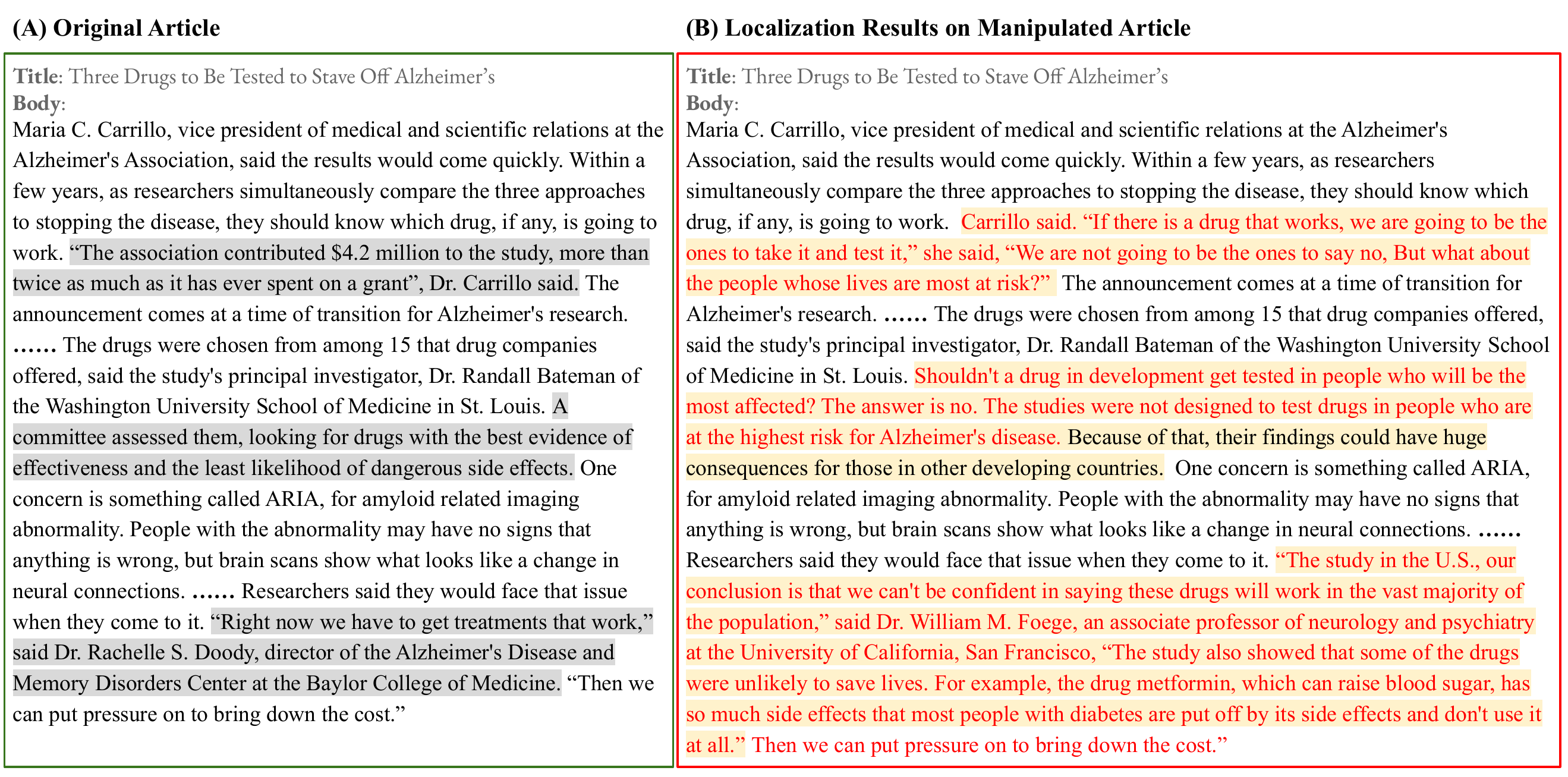}
    \caption{\textbf{A Qualitative Example on GoodNews.} We omit several human-written sections to fit the figure size. The machine-generated sentences are highlighted in light yellow and their original human-written sentences are highlighted in gray. Sentences localized by AdaLoc are marked by red color. We see that Roberta+AdaLoc effectively captures the manipulated segments in the article. See \Cref{sec:experiments_goodnews} for detailed discussion.
    }
    \label{fig:goodnews_qualitative}
\end{figure*}

\noindent \textbf{Quantitative Results.} \Cref{table:goodnews_localization} presents the localization results of various models on GoodNews. We observe that both \emph{vote} and \emph{AdaLoc} boost the localization precision. For instance, Roberta-L+vote achieves a 15 mAP increase over single-sentence prediction methods. Incorporating AdaLoc yields an additional 13 mAP improvement over Roberta-L+vote.

We draw several conclusions from the Table. First, MGT localization appears more challenging than MGT detection. \Eg, while Roberta-Large can achieve over 80\% accuracy in binary detection tasks (as per findings in \citealp{zhang2023show} and \citealp{mitchell2023detectgpt}), it only achieves around 50 mAP in our localization task. Second, multi-sentence prediction methods outperform the single-sentence prediction strategy (\eg, 40.8$\rightarrow$55.9 in mAP of base size, 50.7$\rightarrow$65.0 in mAP of large size), demonstrating that the challenge of detecting short texts can be alleviated by predicting multiple sentences together. Third, \emph{AdaLoc} further boosts performance over \emph{vote}, highlighting the importance of granularity in multi-sentence prediction. 

In addition, our analysis reveals that the difficulty of MGT localization increases with the greater scale of LLMs. However, benefiting from direct access to the target language model, metric-based methods manage to maintain consistent performance regardless of the model scale. 
\smallskip

\noindent \textbf{Qualitative Results.}
\Cref{fig:goodnews_qualitative} presents an article example from GoodNews. From the Figure, we see that the primary messages conveyed by the article can be substantially altered with just a few manipulated sentences (noting the contrast between the original sentences in gray and the machine-generated sentences in light yellow), emphasizing the importance of MGT localization. We observe that while sentences in boundary may occasionally be misidentified (\eg, ``Becaused of that, their findings .... countries.'' is a machine-generated sentence but was misidentified as human-written), AdaLoc can accurately localize the majority of the machine-generated text segments (marked by red). These findings demonstrate that a reliable MGT localization approach can help people defend against misinformation in manipulated articles.
\smallskip

\begin{table*}[t]
\centering
\begin{tabular}{l ccccccc}
\toprule
\textbf{Model} & GPT-2 & GPT-Neo & OPT & GPT-J & GPT-NeoX & \textbf{mAP} & \textbf{All} \\
Scale & -1.5B & -2.7B & -2.7B & -6B & -20B \\
\midrule
\multicolumn{7}{c}{\textbf{(A) AP on VisualNews~\cite{liu2020visualnews}} }\\
\midrule
Random & 16.70 & 16.95 & 16.53 & 17.13 & 16.80 & 16.83 & 16.73 \\
All 0/1 & 16.91 & 16.71 & 16.73 & 17.36 & 16.71 & 16.88 & 16.81 \\
DetectGPT & 37.13 & 37.51 & 35.89 & 35.62 & 36.82 & 36.59 & 36.93 \\
DetectLLM & 38.69 & 37.73 & 39.91 & 38.57 & 38.26 & 38.63 & 38.38 \\
ChatGPT-D & 25.05 & 23.21 & 22.59 & 23.13 & 21.48 & 23.09 & 22.95 \\
Roberta-B & 36.93 & 36.87 & 33.27 & 31.33 & 27.50 & 33.18 & 33.01 \\
Roberta-L & 49.18 & 51.35 & 41.71 & 39.86 & 32.85 & 42.99 & 43.06 \\
\hline
Roberta-B+vote & 53.84 & 53.39 & 47.63 & 46.47 & 39.47 & 48.16 & 48.36 \\
Roberta-L+vote & 66.79 & 66.62 & 58.59 & 56.74 & 47.55 & 59.26 & 59.69 \\
Roberta-L+AdaLoc & \textbf{78.40} & \textbf{78.29} & \textbf{72.37} & \textbf{70.96} & \textbf{64.46} & \textbf{72.90} & \textbf{73.35} \\
\midrule
\multicolumn{7}{c}{\textbf{(B) AP on WikiText~\cite{merity2016pointer}} }\\
\midrule
Random Guess & 15.38 & 14.19 & 14.49 & 13.23 & 14.25 & 14.31 & 14.02 \\
All 0/1 & 14.60 & 13.99 & 14.33 & 13.08 & 13.47 & 13.89 & 13.87 \\
Roberta-B & 36.63 & 32.31 & 30.90 & 23.80 & 21.18 & 28.96 & 29.00 \\
Roberta-L & 45.39 & 40.98 & 35.42 & 28.93 & 23.67 & 34.88 & 34.98\\
\hline
Roberta-B+vote & 51.56 & 43.61 & 40.67 & 30.29 & 27.48 & 38.72 & 39.40 \\
Roberta-L+vote & 64.69 & 58.26 & 49.47 & 40.60 & 33.99 & 49.40 & 50.68 \\
Roberta-L+AdaLoc & \textbf{74.55} & \textbf{70.89} & \textbf{66.83} & \textbf{57.53} & \textbf{54.70} & \textbf{64.90} & \textbf{66.03} \\
\bottomrule
\end{tabular}
\caption{\textbf{Zero-shot Localization Results on VisualNews and WikiText.} Despite being fine-tuned only on GoodNews articles, Roberta-L+AdaLoc boosts performance over Roberta-L+vote on both VisualNews and WikiText, achieving 13.7\% and 15.5\% mAP increases. These gains indicate that AdaLoc is able to identify LLM-generated sentences without overfitting to specific human-written styles in GoodNews. See \Cref{sec:experiments_visualnews_wikitext} for detailed discussion.} 
\vspace{-2mm}
\label{table:zero_shot_localization}
\end{table*}

\begin{table}[t]
\centering
\begin{tabular}{l cccc}
\toprule
\textbf{Size} & Segs=1 & Segs=2 & Segs=3 & \textbf{Avg.} \\
\midrule
$m$=1 & 55.89 & 57.19 & 58.62 & 57.23 \\
$m$=2 & 74.03 & 70.12 & \textbf{66.29} & 70.15 \\
$m$=3 & 78.40 & \textbf{70.59} & 64.11 & \textbf{71.03} \\
$m$=4 & \textbf{78.42} & 66.49 & 59.85 & 68.25 \\
$m$=5 & 76.46 & 61.57 & 55.99 & 64.67\\
\bottomrule
\end{tabular}
\caption{\textbf{Ablation Study of Window Size on GoodNews.} $m$ denotes the number of sentences in a sliding window, ``Segs'' denotes the number of machine-generated segments in an article. With our data generation method, greater number of segments results in shorter text length per segment. We observe that greater $m$ leads to improved performance on segments of long texts, while reduced precision on segments with short texts. See \Cref{sec:experiments_goodnews} for detailed discussion.} 
\label{table:ablation_window_size}
\end{table}

\begin{table}[t]
\centering
\begin{tabular}{l cc}
\toprule
\textbf{Model} & \textbf{mAP} & \textbf{All} \\
\midrule
Roberta-L & 50.67 & 50.85 \\
Roberta-L+vote & 65.02 & 65.50 \\
Roberta-L+AdaLoc(skip) & 67.59 & 67.63 \\
Roberta-L+AdaLoc(middle) & 70.54 & 70.63 \\
Roberta-L+AdaLoc  & \textbf{78.49} & \textbf{79.13} \\
\bottomrule
\end{tabular}
\caption{\textbf{Ablation Study of Vote Strategy on GoodNews.} ``skip'' denotes that there is no overlapping between different chunks, \ie, window step equals three sentences. ``middle'' means we leverage AdaLoc to predict whether the sentence in the middle is machine-generated. By default, AdaLoc is combined with the majority vote strategy. See \Cref{sec:experiments_goodnews} for discussion.}
\label{table:ablation_vote}
\end{table}

\noindent \textbf{Ablation Study on Window Size.} As discussed in \Cref{sec:MGTL_methods}, our multi-sentence prediction algorithms need to find a balance between window size and granularity. \Cref{table:ablation_window_size} provides an ablation study on window size vs. the number of segments. In our data generation process, greater number of segments leads to shorter individual segment lengths. Specifically, Segs = 1, 2, and 3 corresponds to average segment lengths of 168.8, 84.2, and 57.7 tokens, respectively. 

We observe that larger window sizes typically performs better on longer segments, with a reduced precision on shorter segments. $m$ ranging from 2 to 4 correspond to the optimal performance for segment numbers ranging from 1 to 3. Based on the average values across different segment numbers, we set $m$ to 3, \ie, our \emph{vote} and \emph{AdaLoc} methods predict three sentences at once within a window. 
\smallskip

\begin{table*}[th!]
\centering
\setlength{\tabcolsep}{2.2pt}
\begin{tabular}{l ccccc | ccccc}
\toprule
 \multirow{2}{*}{\textbf{Method}} & \multicolumn{5}{c}{\textbf{Essay~\cite{verma2023ghostbuster}}} & \multicolumn{5}{c}{\textbf{WP~\cite{verma2023ghostbuster}}} \\
 & GLM & GPT3.5 & GPT3.5t & GPT4All & \textbf{mAP} & GLM & GPT3.5 & GPT3.5t & GPT4All & \textbf{mAP} \\
\midrule
Random & 20.89 & 27.31 & 49.02 & 24.76 & 30.50 & 19.99 & 24.65 & 42.33 & 20.96 & 26.98 \\
All 0/1 & 20.71 & 27.26 & 49.12 & 24.66 & 30.44 & 20.06 & 24.63 & 42.05 & 21.09 & 26.96 \\
Roberta-L & 42.27 & 33.10 & 47.70 & 38.07 & 40.28 & 56.87 & 29.18 & 32.35 & 36.76 & 38.80 \\
Chat-D & 64.47 & 54.94 & 68.43 & 55.50 & 60.84 & 53.53 & 43.17 & 51.48 & 48.75 & 49.23 \\
Roberta-MPU & \textbf{70.26} & 60.28 & 67.37 & 60.26 & 64.54 & 60.18 & 45.25 & \textbf{60.81} & 54.43 & 55.17 \\
\hline
Roberta-L+vote & 55.46 & 45.62 & 49.32 & 50.27 & 50.17 & \textbf{70.22} & 41.88 & 33.14 & \textbf{62.05} & 51.82 \\
Chat-D+vote & 62.16 & 59.15 & 75.17 & 58.87 & 63.84 & 57.63 & 44.79 & 58.09 & 54.15 & 53.67 \\
Chat-D+AdaLoc & 67.49 & \textbf{68.73} & \textbf{78.53} & \textbf{61.04} & \textbf{68.95} & 61.83 & \textbf{51.33} & 60.58 & 56.38 & \textbf{57.53} \\
\bottomrule
\end{tabular}
\caption{\textbf{Text Localization Results on Essay and WP~\cite{verma2023ghostbuster}.} \emph{AdaLoc} boosts the performance based on ChatGPT-D, demonstrating its generalization ability across different detectors in diverse domains. See \Cref{sec:experiments_Essay_WP} for discussion.} 
\label{table:Essay_WP_localization}
\end{table*}

\noindent \textbf{Ablation Study on Vote Strategy.} 
\Cref{table:ablation_vote} provides ablation studies for the vote strategy within AdaLoc. We observe that AdaLoc, when combined with the majority vote strategy, achieves the best performance. Alternative strategies, such as ``skip'' and ``middle'' achieves lower performance. Detailed ablation results are presented in \Cref{sec:appendix_ablation}.

\subsection{Zero-shot Experiments on VisualNews and WikiText}
\label{sec:experiments_visualnews_wikitext}

Experimental results in \Cref{sec:experiments_goodnews} on GoodNews show that Roberta+AdaLoc outperforms baselines when evaluated on in-domain data, where the training and test sets have similar data distributions. To verify our model is not simply overfitting to GoodNews, we perform zero-shot experiments on VisualNews and WikiText articles, as presented in \Cref{table:zero_shot_localization}. In zero-shot evaluations, Roberta-L+vote and Roberta-L+AdaLoc continue to boost base models in mAP (43.1$\rightarrow$59.7$\rightarrow$73.4 on VisualNews and 34.9$\rightarrow$50.7$\rightarrow$66.0 on WikiText). It illustrates that the improvements offered by \emph{vote} and \emph{AdaLoc} are due to their effectiveness in indentifying LLM-generated text, rather than recognizing the specific human-written styles found in GoodNews articles.

\subsection{MGT Localization on Essay and WP}
\label{sec:experiments_Essay_WP}
In previous sections, we primarily focused on MGT localization in long articles, such as news reports and Wikipedia. To evaluate our approach in diverse domains, we extend our discussion to student essays (Essay) and creative writing (WP) datasets~\cite{verma2023ghostbuster}. Following~\citet{verma2023ghostbuster}, our experiments focuses on detecting text generated by ChatGLM~\cite{du2022glm}, ChatGPT3.5~\cite{ouyang2022training}, ChatGPT3.5-turbo~\cite{ouyang2022training}, and GPT4All~\cite{gpt4all}. \Cref{table:Essay_WP_localization} presents the localization results of different models. Since the training data of ChatGPT-D (Chat-D)~\cite{guo2023close} already includes a substantial number of ChatGPT-generated content, we utilize Chat-D as our backbone to extract chunk features.

The Table illustrates that our methods, \emph{vote} and \emph{AdaLoc} continue to improve the base model's localization performance, verifying that our method's adaptability to various detectors. To validate the effectiveness of context, we further compare Chat-D+AdaLoc to Roberta-MPU~\cite{tian2024multiscale}, a concurrent method for single sentence prediction. From the results, Chat-D+AdaLoc outperforms Roberta-MPU by 2-4 mAP. Roberta-MPU achieves comparable or slightly better results than Chat-D+vote (1-2 mAP boost). We note that Roberta-MPU requires training/finetuning the Roberta architecture on corpora with various lengths of text. In contrast, our Chat-D+vote achieves comparable results to Roberta-MPU by context and does not require any additional training.

\begin{table}[t]
\centering
\begin{tabular}{l cc}
\toprule
\textbf{Method} & Precision & Recall\\
\midrule
GPT-Zero & 64.29 & 18.92 \\
ChatGPT+AdaLoc & \textbf{84.15} & \textbf{40.47} \\
\bottomrule
\end{tabular}
\caption{\textbf{Comparison to GPT-Zero~\cite{tian2023gptzero} on sentence-level analysis.} In our experiments, we observe that GPT-Zero often classifies manipulated articles as human-written, primarily because these articles contain significant portions of human-written text. As a results, GPT-Zero tends to label all sentences in these documents as human-written and fails to identify machine-generated sentences. See \Cref{sec:experiments_gptzero} for detailed discussion.} 
\label{table:gpt-zero}
\end{table}

\begin{table*}[t]
\centering
\setlength{\tabcolsep}{2.5pt}
\begin{tabular}{rl ccccc}
\toprule
 & \textbf{Method} & ChatGLM & GPT3.5 & GPT3.5-t & GPT4All & mAP \\
\midrule
\textbf{(A)}& Roberta-L+vote & 55.46 & 45.62 & 49.32 & 50.27 & 50.17\\
& ChatGPT-D+vote & 62.16 & 59.15 & 75.17 & 58.87 & 63.84 \\
& Roberta-L+AdaLoc (VisualNews) & 63.05 & 52.29 & 51.02 & 59.65 & 56.50\\
& ChatGPT-D+AdaLoc & \textbf{67.49} & \textbf{68.73} & \textbf{78.53} & \textbf{61.04} & \textbf{68.95} \\
\midrule
\textbf{(B)} & Roberta-L+vote & 70.22 & 41.88 & 33.14 & 62.05 & 51.82\\
& ChatGPT-D+vote & 57.63 & 44.79 & 58.09 & 54.15 & 53.67 \\
& Roberta-L+AdaLoc (VisualNews) & \textbf{75.97} & 50.59 & 34.48 & \textbf{65.11} & 56.54 \\
& ChatGPT-D+AdaLoc & 61.83 & \textbf{51.33} & \textbf{60.58} & 56.38 & \textbf{57.53} \\

\bottomrule
\end{tabular}
\caption{\textbf{Out-of-domain Evaluation on Various Document Categories.} Roberta-L+AdaLoc (VisualNews) is finetuned only on VisualNews articles and then evaluated on \textbf{(A)} Essay and \textbf{(B)} WP. We observe that Roberta-L+AdaLoc(VisualNews) achieves 5$\sim$6 mAP boost compared to Roberta-L+vote, validating the generalization ability of AdaLoc. See \Cref{sec:experiments_outofdomain} for discussion.} 
\label{table:outofdomain}
\end{table*}

\subsection{Comparison to GPTZero}
\label{sec:experiments_gptzero}
In our experiments, we note that the online tool GPTZero~\cite{tian2023gptzero} can provide an analysis about which sentences are likely to be AI-generated, which can be used for our localization task. Therefore, we include an additional comparison to GPT-Zero in this Section. Specifically, we randomly select 30 ChatGPT-manipulated articles from the Essay dataset and apply GPT-Zero to analyze which sentences are machine-generated. 

\Cref{table:gpt-zero} reports the comparison between GPTZero and ChatGPT-D+AdaLoc. We observe that GPTZero struggles to retrieve machine-generated sentences in manipulated articles. This is because these articles contain only a few portions generated by LLMs, and are likely to be classified as human-written by GPTZero. In these cases, all sentences in such documents are categorized to human-written, leading to low recall scores. We provide a specific example in \Cref{sec:appendix_gptzero_interface}. We also tested 30 GPT2-manipulated articles on GoodNews, finding that GPTZero encounters the domain shift issue. That is, all GPT2-manipulated articles are classified as human-written.

\subsection{Out-of-domain Evaluation}
\label{sec:experiments_outofdomain}
In \Cref{sec:experiments_visualnews_wikitext}, we validated through zero-shot experiments that our method is capable of recognizing machine-generated sentences in news articles from various sources. To further evaluate AdaLoc's generalization across different types of documents (\eg, news articles, study essays, and creative writings), we conducted out-of-domain experiments on Essay and WP. Specifically, we first fine-tuned Roberta-L+AdaLoc using VisualNews articles, and then directly applied Roberta-L+AdaLoc (VisualNews) to Essay and WP. \Cref{table:outofdomain} presents the experiment results. From the table, we observe that Roberta-L+AdaLoc (VisualNews) outperforms Roberta-L+vote, achieving a 5$\sim$6 mAP improvement. Since our adaptor is fine-tuned only on news articles, the improvements indicate that AdaLoc has a strong generalization ability, even across different document categories including news articles, student essays, and creative writings. This conclusion is also consistent with our zero-shot experiment results in~\Cref{sec:experiments_visualnews_wikitext}.

\section{Conclusion}
\label{sec:conclusion}
In this paper, we conduct a comprehensive study of MGT localization, aiming at recognizing AI-generated sentences within a document. We identify a major challenge in MGT localization as the short spans of text at the sentence level, \ie, single sentence may not provide sufficient information for distinguishing machine-generated content. To address this, we propose our methods, \emph{vote} and \emph{AdaLoc}, to predict multiple sentences together, allowing changes in style or content to boost performance. Our methods are evaluated on five diverse datasets (GoodNews, VisualNews, WikiText, Essay, and WP), achieving a gain of 4$\sim$13\% mAP over baselines. The improvements across various datasets and detectors demonstrate the effectiveness and generalization of our method in MGT localization. 
\medskip

\noindent \textbf{Acknowledgements}
This material is based upon work supported, in part, by DARPA under agreement number HR00112020054. Any opinions, findings, and conclusions or recommendations expressed in this material are those of the author(s) and do not necessarily reflect the views of the supporting agencies.


\clearpage
\section*{Limitations}
In this paper, we highlight short text detection as a primary challenge in MGT localization and introduce our methods, \emph{vote} and \emph{AdaLoc}, to enhance the efficacy of existing detectors. Despite notable improvements across diverse datasets, our findings indicate substantial potential for further enhancement localization metrics, particularly in identifying sentences produced by more advanced language models. For example, ChatGPT+AdaLoc achieves only 69\% and 57\% mAP on Essay and WP, respectively. Roberta-L+AdaLoc obtains 65$\sim$78\% AP for texts generated by GPT-NeoX. Therefore, detection of short texts remains a challenge to be further explored. 

Another challenge in our experiments is the domain-shift issue, where detectors optimized for one domain often experience varying degrees of performance degradation on out-of-domain data. For example, in our experiments, detectors all achieve lower scores on VisualNews and WikiText compared to GoodNews. Thus, enhancing model's generalization across different domains, such as combining GhostBuster~\cite{verma2023ghostbuster} with our method, could be a potential direction for further work. 

In addition, our method is specifically developed to localize the text that is generated by machines directly. Instances where human-written text is paraphrased by LLMs or vice versa are not considered in our study. Therefore, exploring approaches to identify paraphrased content within articles, such as \citet{krishna2024paraphrasing}, represents another area for further work.

\section*{Ethics Statement}
In our study, we propose MGT localization methods (\eg, Roberta-L+AdaLoc or ChatGPT-D+AdaLoc) to identify LLM-generated sentences within text documents, which can be helpful in defending against misinformation spread by LLMs. However, like all other detectors, our system will not produce 100\% accurate predictions, especially when detecting texts from models unseen during training, as well as text domains that are far from the training corpus. Therefore, we strongly discourage incorporating our methods into automatic detection systems without human supervision, such as plagiarism detection or other situations involving suspected use of LLM-generated text. A more suitable application case would be using our methods under human supervision, detecting misinformation generated by LLMs in articles or social media content. We also recognize that bad actors could manipulate articles and spread misinformation according to the data preparation pipeline presented in our paper. Thus, we aim for our paper to highlight the need for building tools like Roberta+AdaLoc to identify and localize manipulated content in such articles.

\bibliography{anthology,custom}

\appendix

\section{Implementation Details}
\label{sec:appendix_implementation_details}
In our experiments, we utilized the open-source tookkit, MGTBench~\cite{he2023mgtbench}, to evaluate various baselines, such as DetectGPT and DetectLLM. Our model is primarily developed based on Pytorch~\cite{paszke2019pytorch} and Transformers~\cite{wolf2020transformers} libraries. We configured our detector to a maximum sequence length of 512 tokens. For Roberta+AdaLoc and ChatGPT-D+AdaLoc, we adopted a batch size of 16 and set the learning rate to $1\times 10^{-5}$. Though increasing the number of training epochs can result in better performance, we finetuned AdaLoc for three epochs with an early stopping mechanism to prevent overfitting to specific data domains, following~\citet{verma2023ghostbuster}. Our experiments were conducted on NVIDIA RTX-A6000 or A40 GPUs, fitting a 48 GB GPU memory requirement. It takes approximate one hour to finetune AdaLoc on chunks extracted from 10,000 manipulated articles.

\section{Additional Results}
\label{sec:appendix_additional_results}

\subsection{Visualization of Manipulated Articles}
\label{sec:appendix_visualization}

\begin{figure*}[t]
    \centering
    \includegraphics[width=.99\linewidth]{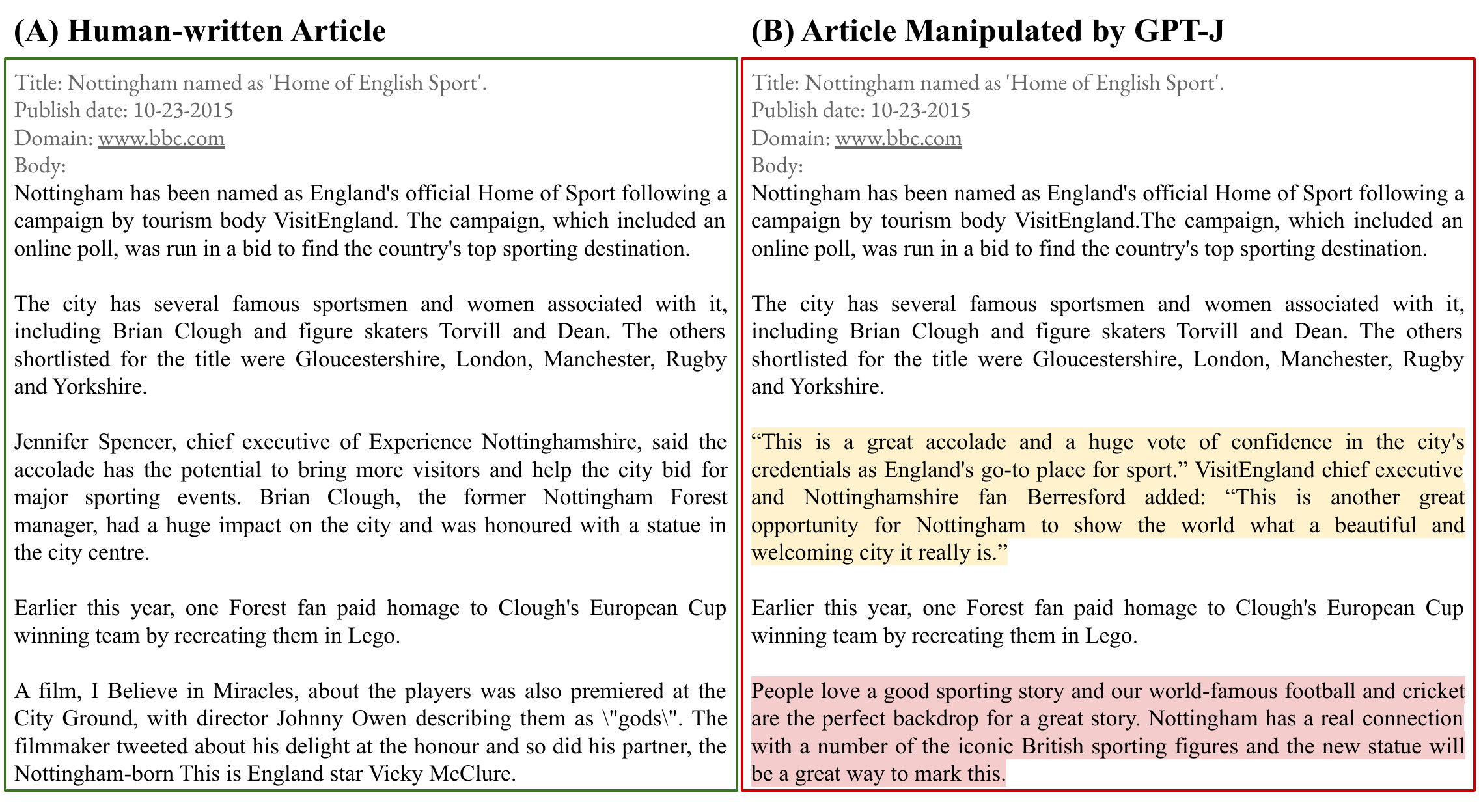}
    \caption{\textbf{A synthesized article consists of both human-written and machine-generated texts.} (A): presents the original human-written article from VisualNews~\cite{liu2020visualnews}; (B): an synthesized article manipulated by GPT-J~\cite{gpt-j}. In this example, we replaced two paragraphs of the original article with machine-generated content, highlighted in yellow and pink. See \Cref{sec:appendix_visualization} for discussion.
    }
    \label{fig:appendix_article_sample}
\end{figure*}

As outlined in \Cref{sec:MGTL_data_preparation}, we use language models to produce sentences and blend these generated sentences with human-written text to create our training and evaluation data. \Cref{fig:appendix_article_sample} provides a specific example of this process as a supplementary example to the main paper. In this example, we replaced two segments of human-written text with content generated by GPT-J~\cite{gpt-j}, highlighted in light yellow and pink, respectively.

\subsection{Comparison to SeqXGPT}
\label{sec:appendix_seqxgpt}
\begin{figure*}[t]
    \centering
    \includegraphics[width=.99\linewidth]{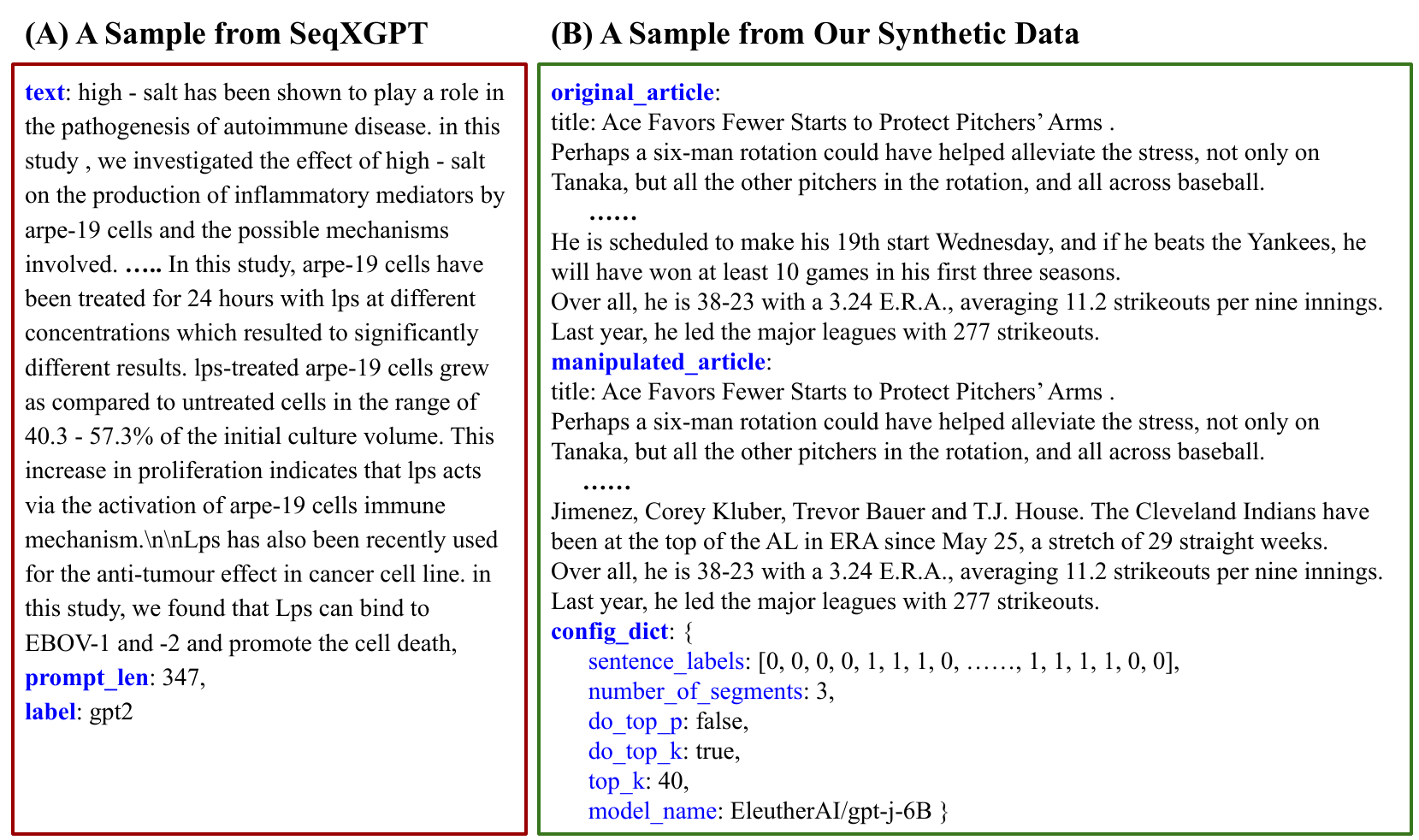}
    \caption{\textbf{Comparison to SeqXGPT.} (A): A sample from SeqXGPT~\cite{wang2023seqxgpt}; (B): A sample generated for our MGT localization task. We observe that SeqXGPT primarily focuses on scenarios where the first part of an article is human-authored, and all subsequent sections are generated by LLMs. In contrast, our approach handles a more realistic and complex scenario, where multiple segments in an article are generated by LLMs. Additionally, our synthetic data provides more fine-grained annotations, including sentence-level labels, LLM sampling strategy, number of machine-generated segments, among others. See \Cref{sec:appendix_seqxgpt} for discussion.
    }
    \label{fig:appendix_seqxgpt_comparison}
\end{figure*}

To demonstrate the differences between our MGT localiztion task and sentence prediction in SeqXGPT~\cite{wang2023seqxgpt}, we provide a specific example in \Cref{fig:appendix_seqxgpt_comparison}. The Figure illustrates that SeqXGPT follows an assumption that articles are structured with an initial human-written segment followed by an AI-generated segment. In constrast, our task involves articles containing multiple machine-generated sections, aligning more closely with real-world application scenarios. In addition, our synthetic data provides more abundant annotations, including sentence-level labels, LLM sampling strategies, and the number of machine-generated segments.

\subsection{Article Assessment Interface of GPT-Zero}
\label{sec:appendix_gptzero_interface}
\begin{figure*}[t]
    \centering
    \includegraphics[width=.75\linewidth]{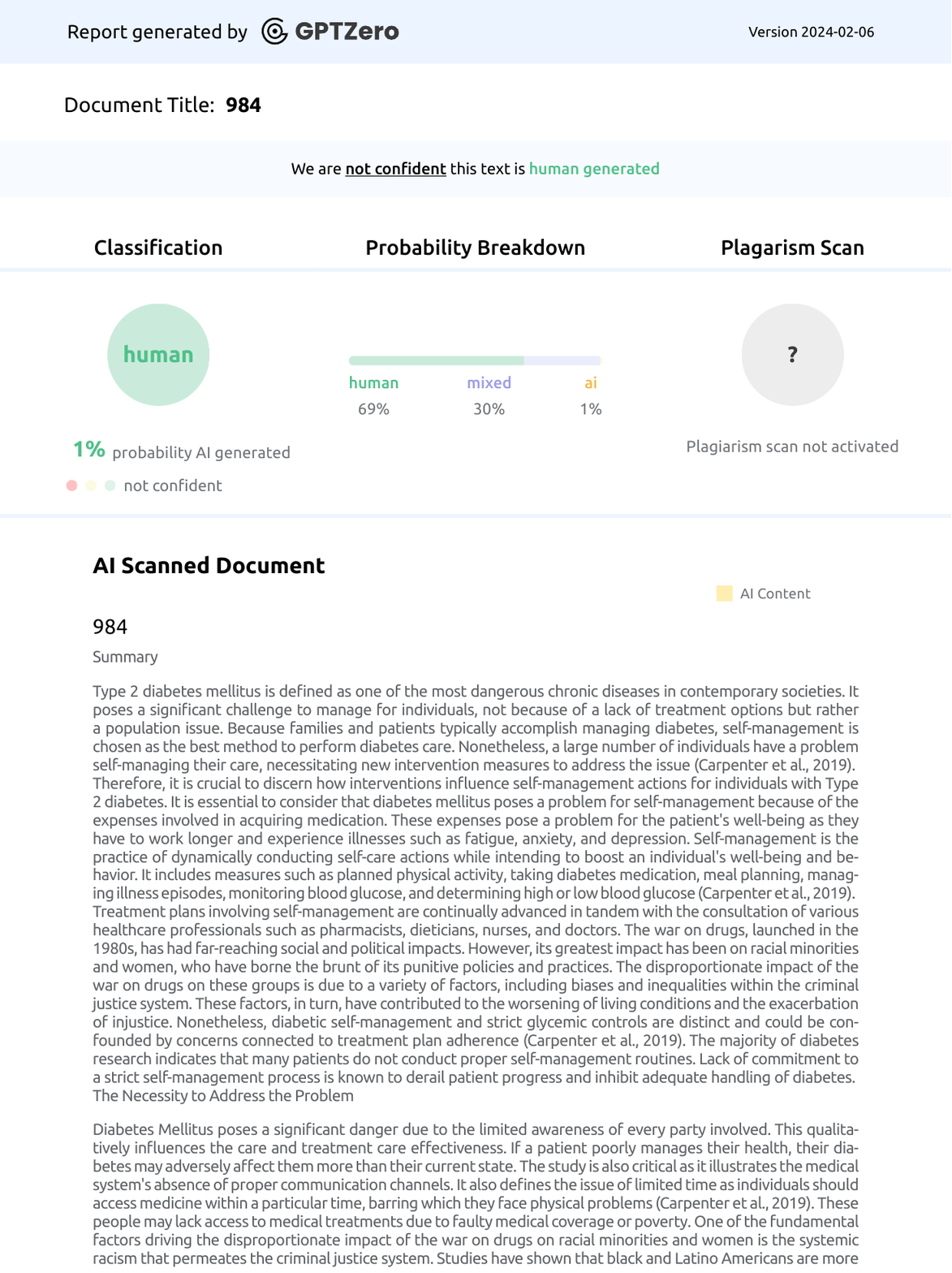}
    \vspace{-6mm}
    \caption{\textbf{An article analyzed by GPT-Zero.} GPT-Zero inaccurately identified the manipulated article as human-generated and consequently label all sentences as human-written. See \Cref{sec:appendix_gptzero_interface} for details.
    }
    \label{fig:appendix_gptzero_sample}
\end{figure*}

A screenshot of an article evaluated by GPT-Zero is provided in \Cref{fig:appendix_gptzero_sample}, supplementing our main paper. From the Figure, we see that GPT-Zero incorrectly identifies the manipulated article as human-written, labeling all sentences within as human-written, which results in low recall scores. This instance supports our discussion in \Cref{sec:experiments_gptzero}. 

\subsection{Ablation Study}
\label{sec:appendix_ablation}
In \Cref{sec:experiments_goodnews}, we mainly ablate the window size, \ie, predicting how many sentences at once within a window. \Cref{table:appendix_ablation} provides additional ablation studies of AdaLoc. We see that AdaLoc, when combined with the majority vote strategy, achieves the best performance. Alternative strategies, such as ``skip'' and ``middle'' achieves lower performance.

\clearpage

\begin{table*}[t]
\centering
\begin{tabular}{l ccccccc}
\toprule
\textbf{Model} & GPT-2 & GPT-Neo & OPT & GPT-J & GPT-NeoX & \textbf{mAP} & \textbf{All} \\
Scale & -1.5B & -2.7B & -2.7B & -6B & -20B \\
\midrule
Roberta-L & 57.24 & 58.05 & 49.38 & 47.41 & 41.32 & 50.67 & 50.85 \\
Roberta-L+vote & 71.03 & 72.03 & 64.37 & 62.28 & 55.39 & 65.02 & 65.50 \\
Roberta-L+AdaLoc(skip) & 72.07 & 72.01 & 67.64 & 65.79 & 60.43 & 67.59 & 67.63 \\
Roberta-L+AdaLoc(middle) & 74.92 & 74.84 & 69.85 & 68.87 & 64.21 & 70.54 & 70.63 \\
Roberta-L+AdaLoc & \textbf{82.82} & \textbf{82.46} & \textbf{78.69} & \textbf{76.90} & \textbf{71.62} & \textbf{78.49} & \textbf{79.13} \\

\bottomrule
\end{tabular}
\caption{\textbf{Ablation Study on GoodNews~\cite{biten2019good}.} ``skip'' denotes that there is no overlapping between different chunks. ``middle'' means we leverage AdaLoc to predict whether the sentence in the middle is machine-generated. By default, AdaLoc is combined with our majority vote strategy. See \Cref{sec:appendix_ablation} for discussion.}
\label{table:appendix_ablation}
\end{table*}



\end{document}